\documentclass[conference,10pt,letterpaper]{IEEEtran}
\usepackage{amsmath,amssymb,graphicx,booktabs,array,multirow}
\usepackage{float,balance,colortbl,placeins}
\usepackage{fontspec,unicode-math}
\usepackage{tikz,pgfplots}
\usetikzlibrary{arrows.meta,positioning,fit,backgrounds}
\pgfplotsset{compat=1.18}

\usepackage[hidelinks]{hyperref}
\hypersetup{pdftitle={Near--Far Attention: Region-Aware Sparse Retrieval for Long-Context LLM Decoding},pdfauthor={}}

\definecolor{nfablue}{HTML}{27658A}
\definecolor{nfateal}{HTML}{257D72}
\definecolor{nfaorange}{HTML}{B76A26}
\definecolor{nfagray}{HTML}{56616D}
\definecolor{nfaaccent}{HTML}{C5414E}
\definecolor{nfadense}{HTML}{7950A6}
\definecolor{nfabaseline}{HTML}{88939F}
\definecolor{nfaheat}{HTML}{443469}
\tikzset{box/.style={draw=nfablue,rounded corners=2pt,fill=nfablue!5,align=center,font=\sffamily\footnotesize,inner sep=5pt},flow/.style={-{Latex[length=2mm]},semithick,draw=nfagray},control/.style={flow,dashed,draw=nfaorange}}
\title{Near--Far Attention: Region-Aware Sparse Retrieval\\for Long-Context LLM Decoding}
\author{\IEEEauthorblockN{YuanShuo Wang, Jianchao Tan, Lei Gong\textsuperscript{*}, Teng Wang, Wenqi Lou and Xuehai Zhou}
\IEEEauthorblockA{University of Science and Technology of China\quad(\textsuperscript{*}Corresponding author)\\
wangyuanshuo@mail.ustc.edu.cn, tanjianchaoustc@gmail.com\\
\{leigong0203, wangt635, louwenqi, xhzhou\}@ustc.edu.cn}}
\begin{document}
\raggedbottom
\maketitle
\begin{abstract}
Sparse attention reduces the cost of long-context decoding, but token selection introduces additional scoring and ranking overhead. Moreover, uniform page scoring overlooks differences in relevance between near and far context. We propose Near--Far Attention (NFA), which adapts retrieval to regional attention patterns while keeping candidate scoring lightweight. During prefill, the Lightweight Gate Estimator identifies regions that can be skipped during decoding, adding less than 1\% to dense-prefill time. For the remaining regions, the Calibration-Free Subspace Extractor supplies compact features shared by page-level and token-level scoring without offline calibration. The Distance-Aware Two-Level Selector then scores near and far pages differently and jointly ranks candidate tokens under a shared budget. Evaluations across model families, parameter scales, and GQA/MLA architectures show aggregate task quality comparable to full attention. On Qwen3-8B at 512K context length, NFA achieves up to $5.49\times$ operator speedup and $2.97\times$ whole-model decoding speedup over FlashInfer, including selection overhead.

\end{abstract}
\begin{IEEEkeywords}
Sparse attention, shared representations, regional retrieval, long-context inference
\end{IEEEkeywords}

\section{Introduction}
Large language models (LLMs) are increasingly used for long-document understanding, multi-turn interactions, and complex reasoning, creating a growing demand for long-context processing. During autoregressive decoding, dense attention repeatedly accesses the historical key–value (KV) cache to generate each new token. As the context grows, the resulting computation and memory traffic increase attention latency, making efficient access to historical information a central challenge in long-context inference. To reduce this cost, recent models such as DeepSeek-V4.1-Flash~\cite{deepseekv41} and LongCat-2.0~\cite{longcatlsa} incorporate sparse attention into their architectures.

Sparse attention addresses this challenge by selecting a subset of historical tokens for attention computation~\cite{quest}. However, identifying relevant tokens introduces additional scoring and ranking costs, which can offset the savings from sparse computation~\cite{hisa,jssa,asynctls}. Under a fixed token budget, the scoring rule determines which historical tokens contribute to attention. Mean-based page scoring can dilute isolated high scores, whereas extrema-based scoring can yield loose relevance bounds. The regional coverage differences observed in our profiling motivate using different scoring rules for near and far context.

These limitations motivate adapting both the scope of retrieval and the scoring of candidates.

We propose Near--Far Attention (NFA). The Lightweight Gate Estimator restricts where to search, the Calibration-Free Subspace Extractor provides a compact scoring subspace shared by both selection levels, and the Distance-Aware Two-Level Selector scores near and far pages differently before jointly ranking candidate tokens. Final attention uses full-dimensional representations.
Our main contributions are as follows:

\begin{itemize}
\item \textbf{Lightweight Gate Estimator.}
We derive per-layer retrieval gates from input-specific attention statistics during prefill. The gates remain fixed throughout decoding and eliminate scoring and selection in disabled regions.
\item \textbf{Calibration-Free Subspace Extractor.}
We construct compact query and key representations shared by page-level and token-level scoring, reducing scoring costs without offline calibration.
\item \textbf{Distance-Aware Two-Level Selector.}
We select candidate pages separately from near and far regions using region-specific scoring rules, then jointly rank their tokens under a shared budget. 
\end{itemize}

We evaluate both the accuracy and efficiency of NFA across multiple models and long-context benchmarks. LongBench v1 evaluations cover different model sizes and GQA/MLA architectures, while Qwen3-8B is further evaluated on LongBench v2 and RULER. Performance measurements on Qwen3-8B over 32K--512K contexts show up to $5.49\times$ operator speedup, including token selection, and $2.97\times$ whole-model decoding speedup over FlashInfer.

\section{Background and Motivation}
\subsection{Attention Architectures}
Autoregressive inference first processes the input during prefill, then generates tokens sequentially during decoding. A key--value (KV) cache avoids recomputing historical keys and values. At each decoding step, a query head computes

\begin{equation}
\mathbf{o}
=
\operatorname{softmax}\!\left(
\frac{\mathbf{q}\mathbf{K}^{\top}}{\sqrt{d}}
\right)\mathbf{V}.
\label{eq:attention}
\end{equation}
Here, $\mathbf{q}\in\mathbb{R}^{1\times d}$ is the current query,
$\mathbf{K}\in\mathbb{R}^{L\times d}$ and
$\mathbf{V}\in\mathbb{R}^{L\times d_v}$ contain the keys and values
at the $L$ visible positions, and softmax normalizes over these positions.
Head and layer indices are omitted for clarity.

Grouped-query attention (GQA) reduces KV storage by sharing keys and values within each group of query heads~\cite{gqa}. Multi-head latent attention (MLA) further compresses content keys and values into a shared latent vector, while caching a separate key for rotary position embeddings (RoPE)~\cite{rope}. Its attention logits combine non-RoPE (NoPE) content and RoPE positional terms~\cite{deepseekv2}. Both architectures retain head-specific attention distributions. Although they reduce storage per token, dense attention still accesses every visible position; sparse attention reduces this work by selecting a subset.

\subsection{Sparse Attention for Long-Context Decoding}
Existing methods reduce sparse decoding costs through compact scoring representations, hierarchical candidate filtering, and restricted historical access. These complementary strategies reduce per-candidate scoring cost, the number of candidates, and the scope of historical access, respectively.

\textbf{Compact scoring representations.}
Token selection can use compact features instead of full-dimensional representations. FASA identifies dominant RoPE frequency chunks through offline calibration~\cite{fasa}, whereas JSSA derives a scoring subspace from joint query--key statistics~\cite{jssa}. These representations lower the cost per candidate, but do not by themselves reduce the number of positions scored.

\textbf{Hierarchical candidate filtering.}
Page summaries narrow the search before fine-grained scoring. Quest scores pages using coordinate-wise key extrema and attends to the selected pages~\cite{quest}. ClusterKV instead groups keys into semantic clusters and retrieves relevant clusters to reduce recall overhead~\cite{clusterkv}. AsyncTLS adds token refinement within candidate blocks~\cite{asynctls}, while HISA filters pooled indexer keys before applying its original indexer~\cite{hisa}. LongCat Sparse Attention combines block-to-token filtering with cross-layer index reuse~\cite{longcatlsa}, while DeepSeek-V4.1-Flash restricts later indexers to a candidate pool constructed by an earlier layer~\cite{deepseekv41}. Because refinement cannot recover excluded tokens, coarse selection must preserve relevant candidates under a limited budget.

\textbf{Restricted historical access.}
Historical access can also be restricted according to locality or head behavior. StreamingLLM retains attention sinks and a recent window~\cite{streaming}. DuoAttention uses offline optimization to distinguish retrieval heads requiring full history from streaming heads using sinks and recent tokens~\cite{duoattention}. FastGen instead profiles the input to choose head-specific retention policies~\cite{fastgen}. These designs motivate adapting retrieval scope to attention patterns.

\subsection{Observations and Design Rationale}
\begin{figure}[!tbp]
\centering
\includegraphics[width=\columnwidth]{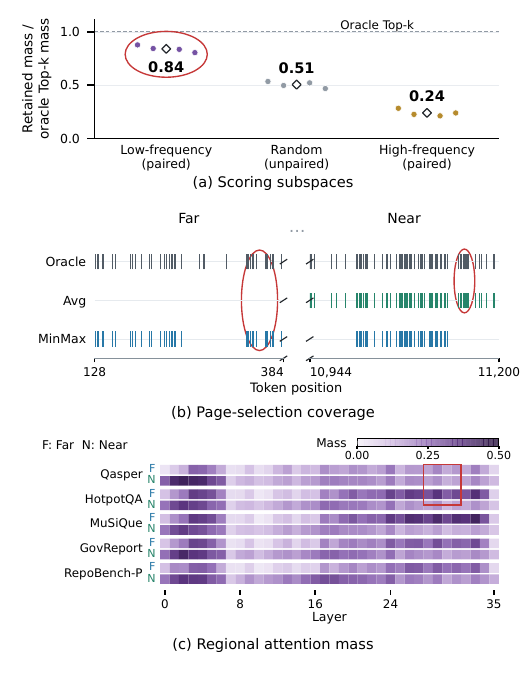}
\caption{Observations on Qwen3-8B. (a) Retained attention mass relative to oracle TopK under a shared 32-coordinate budget; dots denote documents and diamonds denote means. (b) A representative example of the regional coverage differences between Avg and MinMax observed across profiled queries. (c) Regional attention patterns over contiguous query windows, summarized by median attention mass for each task and layer (F: far; N: near). Profiling settings are given in Section~\ref{sec:setup}.}

\label{fig:motivation}
\end{figure}
Three observations guide NFA's representation, selection, and gating designs.

\textbf{Scoring representations.}
Under the same 32-coordinate budget and two-level selector, low-frequency coordinates retain 84\% of oracle TopK attention mass across four documents, versus 51\% for random and 24\% for high-frequency coordinates (Fig.~\ref{fig:motivation}(a)). The oracle uses full-dimensional attention probabilities. These results support using a fixed low-frequency subspace for GQA retrieval.

\textbf{Regional differences.}
Avg scores pages using their mean keys, equivalently averaging token-level dot products, which can dilute isolated high scores. MinMax uses coordinate-wise key extrema to form an upper bound, but these extrema may come from different tokens and overestimate a page’s relevance. Across the profiled queries, Avg generally covers more oracle tokens in the near region, whereas MinMax provides better coverage in the far region. Figure~\ref{fig:motivation}(b) illustrates this trend with a representative query. This motivates region-specific page scoring and separate near- and far-page selection followed by joint token ranking.

\textbf{Regional attention patterns.}
Regional attention patterns vary across tasks and layers. In the profiled inputs, some layers consistently allocate little attention mass to the near or far region from prefill through subsequent decoding; Fig.~\ref{fig:motivation}(c) summarizes these patterns by task and layer. This sustained low regional attention suggests weak layer-level dependence on the corresponding region for the current input, motivating input-specific, per-layer gates that bypass retrieval in low-attention regions during decoding.

\section{Near--Far Attention}
\begin{figure*}[!t]
\centering
\includegraphics[width=\textwidth]{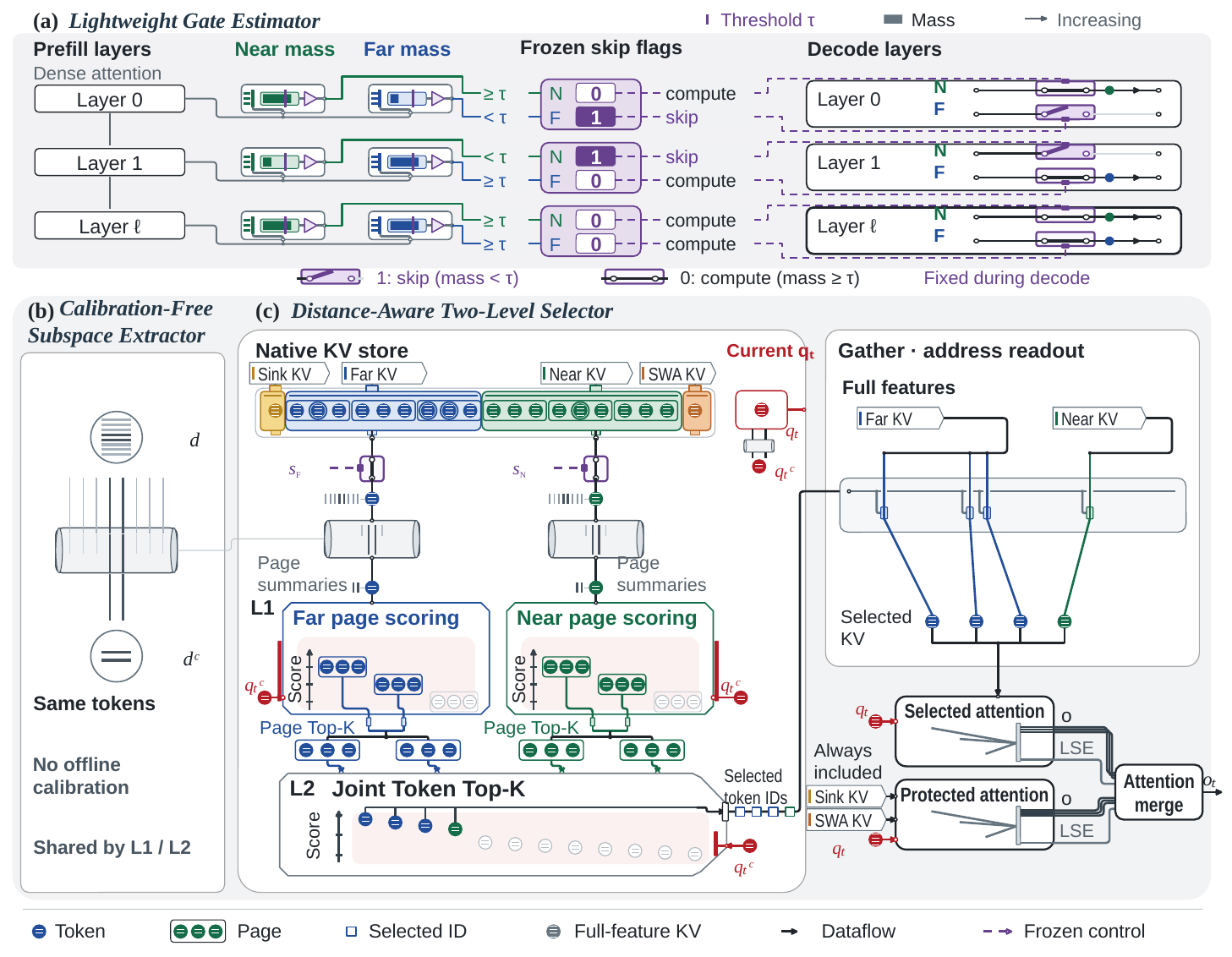}
\caption{Overview of NFA. (a) Prefill-time regional attention statistics determine per-layer skip flags, which remain fixed during decoding (1: skip; 0: compute). (b) Compact query and key representations are shared by both selection levels. (c) Region-specific page selection is followed by joint token ranking; selected and protected positions participate in full-dimensional attention.}
\label{fig:overview}
\end{figure*}
Figure~\ref{fig:overview} shows NFA's three components. During prefill, the Lightweight Gate Estimator (a) fixes per-layer regional retrieval gates. During decoding, the Calibration-Free Subspace Extractor (b) supplies compact features to the Distance-Aware Two-Level Selector (c), which applies distinct near- and far-page scoring within enabled regions and jointly ranks candidate tokens. NFA then computes attention over the selected and protected positions using the original query, key, and value representations.

NFA separates the visible context into protected positions and retrieval candidates.
At decoding position $t$, the protected set $\mathcal P_t$ contains the first $L_{\mathrm{sink}}$ tokens and the most recent $L_{\mathrm{swa}}$ tokens, which always participate in attention.
The remaining visible positions are partitioned into a near region $\mathcal N_t$ and a far region $\mathcal F_t$.
The near region covers a nominal span of $L_{\mathrm{near}}$ tokens immediately preceding the recent window, with its boundary aligned to a page boundary.
The far region contains the remaining older positions.
Regional gates control retrieval within these two regions, while protected positions bypass both selection levels.

\subsection{Lightweight Gate Estimator}
\label{sec:gates}

The Lightweight Gate Estimator (Fig.~\ref{fig:overview}(a)) uses prefill regional attention statistics as a proxy for each layer's decoding retrieval needs. It aggregates near- and far-region attention mass over the last \(L_{\mathrm{obs}}\) input queries to determine per-layer gates.

Let $\mathcal O$ denote the query positions in this
observation window.
At layer $\ell$, let $p_{\ell,h,i,j}$ be the attention
probability from query position $i$ to key position $j$
in head $h$, normalized over all causally visible keys.
We sum these probabilities within each region and
average over the $H_q$ query heads and observation
queries:
\begin{equation}
D_\ell^{R}
=
\frac{1}{|\mathcal O|H_q}
\sum_{i\in\mathcal O}
\sum_{h=1}^{H_q}
\sum_{j\in\mathcal R_i}
p_{\ell,h,i,j},
\qquad R\in\{N,F\},
\label{eq:regional-mass}
\end{equation}
where $\mathcal R_i$ denotes $\mathcal N_i$ or
$\mathcal F_i$ for the corresponding region.

We augment dense-prefill attention with two regional accumulators for each query and head in the observation window. As each key tile is processed, these accumulators reuse the unnormalized attention weights to sum contributions from the near and far regions. Whenever the running maximum changes, the regional accumulators are rescaled by the same factor as the softmax normalizer. After processing all causally visible keys, we normalize the regional sums and average the resulting masses over observation queries and heads to obtain $D_\ell^R$ in Eq.~\eqref{eq:regional-mass}. This fusion avoids both a separate query--key scoring pass and materialization of the attention matrix.

For each input and layer, retrieval in each nonempty
region is independently disabled when its estimated
mass falls below a threshold $\tau$.
These layer-level decisions are shared across all query heads. The regional skip decisions remain fixed throughout decoding, while token selection within enabled regions is updated using each decoding query. Region membership follows the current decoding position. Protected positions always participate in attention.

When both regions are skipped, only protected-set
assembly remains in the selection path.

\subsection{Calibration-Free Subspace Extractor}\label{sec:index}
The extractor (Fig.~\ref{fig:overview}(b)) applies a fixed, architecture-specific coordinate subset to native queries and keys, yielding compact $q^c$ and $k^c$ shared by page and token selection without offline calibration or a learned projection.

For the GQA models considered here, the observation
in Fig.~\ref{fig:motivation}(a) motivates selecting
the $p$ lowest-frequency RoPE pairs.
After applying RoPE, we retain both coordinates of each
selected pair, yielding a compact representation of
dimension $d^c=2p$.
For MLA, experiments on GLM-4.7-Flash using HotpotQA and GovReport show that several 32-coordinate NoPE subsets retain average quality close to full attention (Fig.~\ref{fig:subspace-ablation}(a)). We use the last 32 coordinates of the per-head NoPE queries and keys as a simple fixed default for both selection levels.

\subsection{Distance-Aware Two-Level Selector}
The selector scores near and far pages differently using compact representations, then jointly ranks candidate tokens from enabled regions under a shared budget (Fig.~\ref{fig:overview}(c)).

\textbf{L1: regional page selection.}
L1 uses mean-key scoring for near pages and MinMax
scoring for far pages,
motivated by the regional coverage differences
in Fig.~\ref{fig:motivation}(b).
For each selection unit $u$, the regional page scores are
\begin{align}
S^N_{b,u}
&=(\bar q^c_{t,u})^\top\mu_{b,u},
\nonumber\\
S^F_{b,u}
&=\frac{1}{|\mathcal H_u|}
\sum_{h\in\mathcal H_u}\sum_{r=1}^{d^c}
\left[
(q^c_{t,h,r})^+ U_{b,u,r}
+(q^c_{t,h,r})^- L_{b,u,r}
\right],
\label{eq:page}
\end{align}
A selection unit $u$ corresponds to a GQA KV group
or an MLA query head.
Let $\mathcal H_u$ denote the query heads associated
with unit $u$ (a singleton for MLA), and let
$\bar q^c_{t,u}$ be their mean compact query.
For each page $b$, let $\mu_{b,u}$ be its mean compact
key, and let $L_{b,u,r}$ and $U_{b,u,r}$ be the minimum
and maximum key values along compact coordinate $r$.
Here $d^c$ is the compact dimension,
$x^+=\max(x,0)$, and $x^-=\min(x,0)$.
MinMax chooses the key extremum according to each
query coordinate's sign.
Because this operation is nonlinear in the query,
we compute each head's page score before averaging
across heads.

Because mean-based and extrema-based scores are not
directly comparable, NFA performs page ranking
independently within each enabled region.
The near and far page budgets, $b_N$ and $b_F$,
satisfy $b_N+b_F=K_1$.
Each region contributes up to its assigned number
of highest-scoring pages; unused slots are not
transferred to the other region.
Within enabled regions, up to two partial pages adjacent
to protected positions bypass page ranking and enter L2 directly.
We expand the admitted pages into token positions,
exclude protected and causally invisible positions,
and remove duplicates to form the candidate set
$\mathcal C_{t,u}$ for L2.

\textbf{L2: joint token selection.}
L2 pools the candidates retained by L1 from both regions
and ranks them using a common compact query--key
dot product.
The total attention budget $K_2\geq|\mathcal P_t|$
includes protected positions, leaving
$K_2-|\mathcal P_t|$ slots for retrieved tokens.
We select the highest-scoring candidates jointly
across regions and form the final attention set:
\begin{align}
z_{j,u}
&=(\bar q^c_{t,u})^\top k^c_{j,u},
\qquad j\in\mathcal C_{t,u},
\nonumber\\
k_{t,u}
&=\min\{K_2-|\mathcal P_t|,|\mathcal C_{t,u}|\},
\nonumber\\
\mathcal I_{t,u}
&=\operatorname{TopK}_{j\in\mathcal C_{t,u}}
(z_{j,u},k_{t,u}),
\nonumber\\
\mathcal A_{t,u}
&=\mathcal P_t\cup\mathcal I_{t,u}.
\label{eq:tokens}
\end{align}
The selected indices identify the native keys and values
used for attention. Each query head computes its own
attention distribution over $\mathcal A_{t,u}$ using
the original, full-dimensional representations.
As shown in Fig.~\ref{fig:overview}(c), attention over selected and protected positions is computed separately and combined using their log-sum-exp normalizers, yielding attention over their union. When both regions are skipped, only protected attention is required.
NFA retains the full KV cache and restricts attention
through the selected indices.

\FloatBarrier
\section{Evaluation}
\subsection{Experimental Setup}\label{sec:setup}
\textbf{Models and benchmarks.} We evaluate Qwen3-8B~\cite{qwen3} on LongBench v1~\cite{longbench}, LongBench v2~\cite{longbenchv2}, and RULER~\cite{ruler}. LongBench v1 evaluations extend to Qwen3-32B, Llama-3.1-8B~\cite{llama3}, and GLM-4.7-Flash~\cite{glm47flash}, covering 16 English tasks. LongBench v2 uses 503 direct-answer questions with middle truncation to 31,500 tokens. RULER reports averages over 11 tasks (excluding QA1/QA2), with 100 examples per task at 32K/64K/128K. Llama uses base-model prompts; Llama and GLM inputs are capped at 31,500 tokens.

\textbf{Configurations and baselines.} Unless otherwise specified, NFA uses 64-token pages, $d^c=32$, $K_1=128$ pages (32 near/96 far), $K_2=1024$ total attention tokens (including protected positions), $L_{\mathrm{near}}=4096$, four sink tokens, and a 128-token window. Gates use 256 tail queries and $\tau_N=\tau_F=0.10$. We compare quality against FullKV, Quest~\cite{quest}, JSSA~\cite{jssa}, and AsyncTLS~\cite{asynctls}. Sparse methods use matched total attention budgets of 1024 tokens on LongBench and RULER; JSSA uses rank 32 (64 for GLM), and AsyncTLS uses 128 coarse 64-token blocks. Quest and JSSA quality evaluations use official implementations.

\textbf{Efficiency measurements.} We benchmark BF16 Qwen3-8B on one H20-3e GPU with CUDA Graphs at 32K--512K and batch sizes 1/2/4 where memory permits. FlashInfer v0.5.2~\cite{flashinfer} uses 64-token pages. YaRN~\cite{yarn} factors are 2/4/8/16 at 64K/128K/256K/512K. We report median time per output token (TPOT) after warmup, including selection overhead and excluding prefill and initialization. Operator latency covers selection and attention across all layers, excluding KV-cache updates. 

For latency comparisons, we evaluate NFA, Quest, JSSA, and AsyncTLS within a common inference framework using matched total attention budgets and the same timing protocol.

\textbf{Observation profiling.} Figure~\ref{fig:motivation}(a) uses four documents, layers 8/24, 16 queries per layer, eight KV groups, and oracle $k=\min(768,\text{eligible tokens})$. Panel (b) shows HotpotQA document 6, layer 8, query 12,915 (boundary 8,640), selected from 256 profiled queries. Panel (c) summarizes 83 requests containing both regions.

\subsection{Task Quality}\label{sec:quality}
\begin{figure}[!tbp]
\centering
\begin{tikzpicture}
\begin{axis}[scale only axis,width=2.65cm,height=1.95cm,xmin=39,xmax=54,
xtick={40,45,50},ymin=-.5,ymax=4.5,ytick={0,1,2,3,4},
yticklabels={NFA,Quest,JSSA,AsyncTLS,FullKV},
xlabel={Average score},title={(a) LongBench v1},
tick label style={font=\scriptsize},label style={font=\scriptsize},title style={font=\small},
axis x line*=bottom,axis y line*=left,tick align=outside,xmajorgrids,grid style={gray!16},clip=false]
\addplot[nfadense!60,dashed,no marks] coordinates {(49.39,-.5) (49.39,4.5)};
\addplot[only marks,color=nfadense,mark=o,mark size=1.8pt] coordinates {(49.39,4)};
\node[font=\scriptsize,anchor=west,text=nfadense] at (axis cs:49.69,4) {49.39};
\addplot[only marks,color=nfablue,mark=triangle*,mark size=1.8pt] coordinates {(40.51,3)};
\node[font=\scriptsize,anchor=west,text=nfablue] at (axis cs:40.81,3) {40.51};
\addplot[only marks,color=nfateal,mark=square*,mark size=1.8pt] coordinates {(49.28,2)};
\node[font=\scriptsize,anchor=west,text=nfateal] at (axis cs:49.58,2) {49.28};
\addplot[only marks,color=nfabaseline,mark=diamond*,mark size=1.8pt] coordinates {(45.13,1)};
\node[font=\scriptsize,anchor=west,text=nfabaseline] at (axis cs:45.43,1) {45.13};
\addplot[only marks,color=nfaaccent,mark=star,mark size=2.6pt] coordinates {(49.27,0)};
\node[font=\scriptsize,anchor=west,text=nfaaccent] at (axis cs:49.57,0) {49.27};
\end{axis}
\begin{axis}[at={(3.85cm,0)},anchor=south west,scale only axis,width=2.65cm,height=1.95cm,xmin=31.7,xmax=33.9,
xtick={32,33},ymin=-.5,ymax=4.5,ytick={0,1,2,3,4},yticklabels={},
xlabel={Accuracy (\%)},title={(b) LongBench v2},
tick label style={font=\scriptsize},label style={font=\scriptsize},title style={font=\small},
axis x line*=bottom,axis y line*=left,tick align=outside,xmajorgrids,grid style={gray!16},clip=false]
\addplot[nfadense!60,dashed,no marks] coordinates {(32.21,-.5) (32.21,4.5)};
\addplot[only marks,color=nfadense,mark=o,mark size=1.8pt] coordinates {(32.21,4)};
\node[font=\scriptsize,anchor=west,text=nfadense] at (axis cs:32.27,4) {32.21};
\addplot[only marks,color=nfablue,mark=triangle*,mark size=1.8pt] coordinates {(32.01,3)};
\node[font=\scriptsize,anchor=west,text=nfablue] at (axis cs:32.07,3) {32.01};
\addplot[only marks,color=nfateal,mark=square*,mark size=1.8pt] coordinates {(32.8,2)};
\node[font=\scriptsize,anchor=west,text=nfateal] at (axis cs:32.86,2) {32.80};
\addplot[only marks,color=nfabaseline,mark=diamond*,mark size=1.8pt] coordinates {(32.21,1)};
\node[font=\scriptsize,anchor=west,text=nfabaseline] at (axis cs:32.27,1) {32.21};
\addplot[only marks,color=nfaaccent,mark=star,mark size=2.6pt] coordinates {(32.41,0)};
\node[font=\scriptsize,anchor=west,text=nfaaccent] at (axis cs:32.47,0) {32.41};
\end{axis}
\end{tikzpicture}
\par\smallskip
\begin{tikzpicture}
\begin{axis}[tick label style={font=\scriptsize},label style={font=\scriptsize},title style={font=\small},axis x line*=bottom,axis y line*=left,grid style={gray!16},tick align=outside,mark size=1.8pt,scale only axis,width=6.5cm,height=1.95cm,xmin=-.12,xmax=2.12,ymin=15,ymax=63,xtick={0,1,2},xticklabels={32K,64K,128K},ytick={20,40,60},ylabel={Score},xlabel={Context length},title={(c) RULER},grid=major,legend style={font=\scriptsize,at={(.5,-.55)},anchor=north,draw=none,fill=none,legend columns=5,/tikz/every even column/.append style={column sep=3pt}}]
\addplot[dashed,semithick,color=nfadense,mark=o] coordinates {(0.000000,59.380000) (1.000000,48.540000) (2.000000,46.430000)};
\addlegendentry{FullKV}
\addplot[thick,color=nfabaseline,mark=triangle*] coordinates {(0.000000,42.850000) (1.000000,26.250000) (2.000000,19.330000)};
\addlegendentry{Quest}
\addplot[thick,color=nfateal,mark=square*] coordinates {(0.000000,59.130000) (1.000000,48.400000) (2.000000,48.310000)};
\addlegendentry{JSSA}
\addplot[thick,color=nfablue,mark=diamond*] coordinates {(0.000000,50.090000) (1.000000,34.110000) (2.000000,37.470000)};
\addlegendentry{AsyncTLS}
\addplot[thick,color=nfaaccent,mark=star,mark size=2.6pt] coordinates {(0.000000,59.020000) (1.000000,47.800000) (2.000000,46.820000)};
\addlegendentry{NFA}
\end{axis}
\end{tikzpicture}
\caption{Qwen3-8B task quality: (a) LongBench v1 average; (b) LongBench v2 accuracy; (c) RULER scores across context lengths. Vertical dashed lines in (a) and (b) mark FullKV. Configurations are given in Section~\ref{sec:setup}.}
\label{fig:main-quality}
\end{figure}
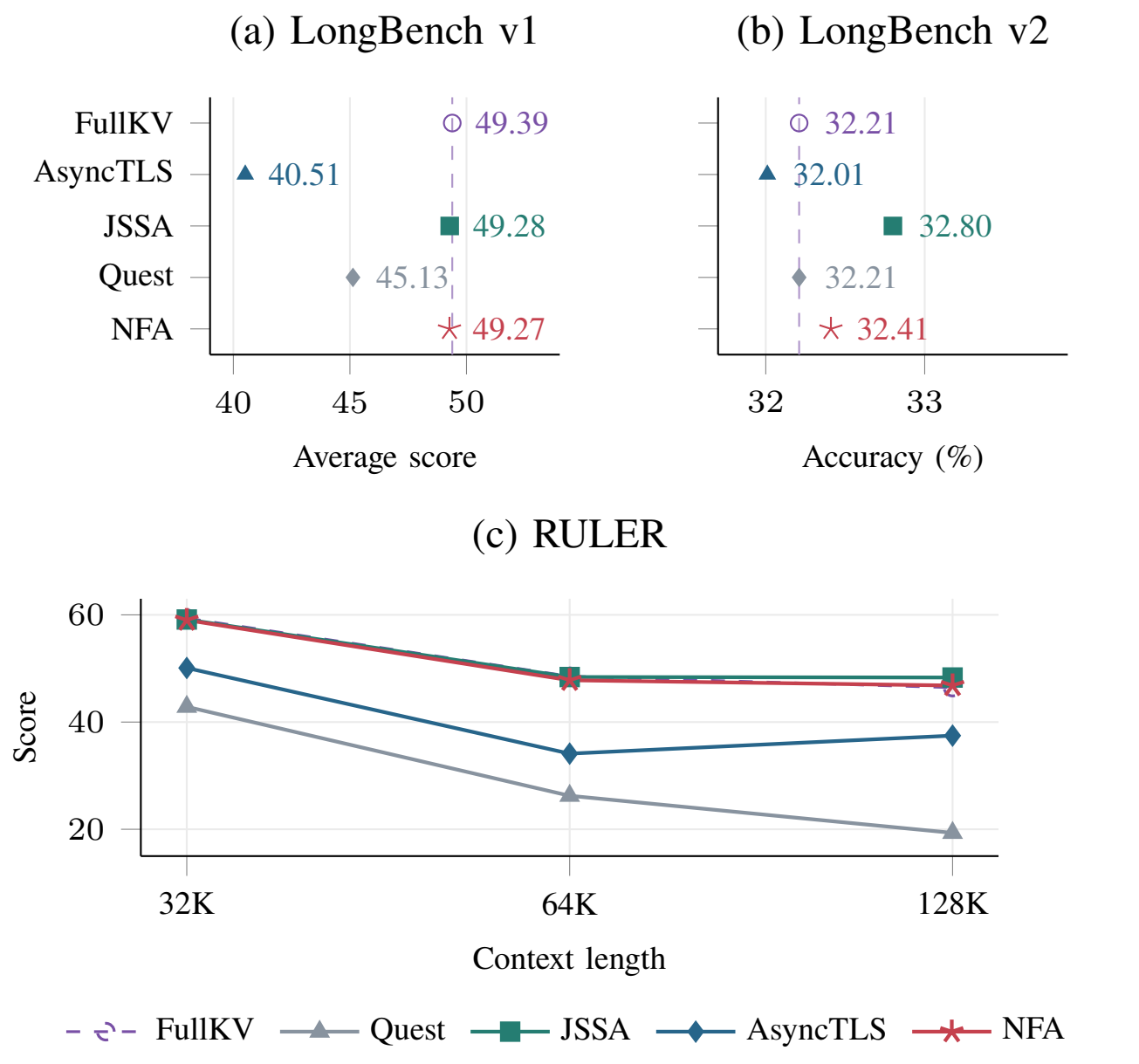
\textbf{Comparison with sparse baselines.} We compare task quality on Qwen3-8B (Fig.~\ref{fig:main-quality}). On LongBench v1, NFA achieves an average score comparable to JSSA and FullKV, while exceeding Quest and AsyncTLS by 4.14 and 8.76 points, respectively (a). It also retains FullKV-level accuracy on LongBench v2 (b). On RULER, NFA outperforms Quest and AsyncTLS across all tested lengths, with margins of 27.49 and 9.35 points at 128K (c).

\textbf{Results across models and architectures.} To assess generality, we extend LongBench v1 evaluation to three additional models (Table~\ref{tab:cross_model}). NFA ranks first among the evaluated sparse methods on Llama-3.1-8B and GLM-4.7-Flash, and second to JSSA on Qwen3-32B. Its averages remain within 0.47 points of FullKV.

\begin{table}[!t]
\centering
\caption{LongBench v1 scores across models. Categories average their constituent tasks; Avg. is the unweighted mean over 16 tasks. Bold and underlined values mark the best and second-best sparse methods; FullKV is the dense reference.}
\label{tab:cross_model}
\begingroup
\footnotesize
\setlength{\tabcolsep}{1.8pt}
\renewcommand{\arraystretch}{1.0}
\begin{tabular*}{\columnwidth}{@{\extracolsep{\fill}}l*{7}{r}@{}}
\toprule
Method & \shortstack{S-Doc\\QA} & \shortstack{M-Doc\\QA} & Summ. & \shortstack{Few-\\shot} & Syn. & Code & Avg.\\
\midrule
\rowcolor{violet!12}
\multicolumn{8}{c}{\textbf{Qwen3-32B (GQA; 16 tasks)}}\\
FullKV & 40.88 & 47.73 & 27.47 & 59.61 & 59.25 & 16.20 & 42.37 \\
Quest & 36.13 & 38.96 & \underline{27.20} & 49.53 & 51.88 & \textbf{26.51} & 38.26 \\
JSSA & \textbf{42.17} & \textbf{50.53} & \textbf{27.44} & \underline{60.48} & \textbf{58.25} & 15.55 & \textbf{43.09} \\
AsyncTLS & 37.53 & 43.36 & 27.02 & 55.01 & 49.16 & \underline{16.95} & 38.81 \\
\cmidrule(lr){1-8}
\textbf{NFA} & \underline{39.97} & \underline{46.83} & 27.19 & \textbf{60.52} & \underline{57.75} & 15.68 & \underline{41.90} \\
\midrule
\rowcolor{nfablue!12}
\multicolumn{8}{c}{\textbf{Llama-3.1-8B base (GQA; 16 tasks)}}\\
FullKV & 20.61 & 15.13 & 19.24 & 69.52 & 52.19 & 68.49 & 38.43 \\
Quest & 16.25 & 9.65 & 15.24 & 66.87 & \underline{22.26} & 63.92 & 31.02 \\
JSSA & \textbf{22.13} & \underline{10.89} & \textbf{19.09} & 68.79 & 19.23 & \textbf{68.75} & \underline{33.67} \\
AsyncTLS & 20.08 & 10.67 & 18.00 & \underline{69.76} & 12.45 & 64.30 & 31.81 \\
\cmidrule(lr){1-8}
\textbf{NFA} & \underline{21.12} & \textbf{14.55} & \underline{19.06} & \textbf{69.83} & \textbf{52.86} & \underline{68.24} & \textbf{38.49} \\
\midrule
\rowcolor{nfateal!12}
\multicolumn{8}{c}{\textbf{GLM-4.7-Flash (MLA; 16 tasks)}}\\
FullKV & 40.94 & 47.20 & 27.86 & 70.92 & 52.50 & 64.91 & 49.72 \\
Quest & 30.03 & 34.84 & 25.96 & 57.46 & 51.18 & 54.02 & 40.95 \\
JSSA & 28.26 & 27.15 & 21.82 & 30.76 & \underline{52.50} & 40.63 & 31.89 \\
AsyncTLS & \underline{40.19} & \textbf{47.71} & \textbf{27.65} & \underline{70.03} & 50.00 & \textbf{64.60} & \underline{49.12} \\
\cmidrule(lr){1-8}
\textbf{NFA} & \textbf{41.32} & \underline{45.68} & \underline{27.54} & \textbf{70.81} & \textbf{52.75} & \underline{63.55} & \textbf{49.29} \\
\bottomrule
\end{tabular*}
\endgroup
\end{table}

\subsection{Decoding Efficiency}

\label{sec:performance}

\begin{figure}[!htbp]
\centering
\begin{tikzpicture}
\begin{axis}[at={(0,2.65cm)},anchor=south west,scale only axis,width=7cm,height=1.65cm,xmin=-.1,xmax=4.1,ymin=0,ymax=5.9,ytick={0,2,4},title={(a) Operator, including selection},xtick={0,1,2,3,4},xticklabels={32K,64K,128K,256K,512K},ylabel={Speedup ($\times$)},title style={font=\small},tick label style={font=\scriptsize},label style={font=\scriptsize},tick align=inside,grid=major,grid style={gray!15},legend style={font=\scriptsize,at={(.02,.98)},anchor=north west,draw=none,fill=white,fill opacity=.85,text opacity=1,row sep=-2pt},mark size=1.6pt]
\addplot[gray,dashed,forget plot] coordinates {(0,1)(4,1)};
\addplot[color=nfablue,mark=*,mark options={solid,fill=nfablue},thick] coordinates {(0,0.460093) (1,0.719343) (2,1.340665) (3,2.511433) (4,5.488573)};\addlegendentry{Batch 1}
\addplot[color=nfateal,mark=square*,mark options={solid,fill=nfateal},thick] coordinates {(0,0.709390) (1,1.221007) (2,2.233095) (3,4.913768)};\addlegendentry{Batch 2}
\addplot[color=nfaorange,mark=triangle*,mark options={solid,fill=nfaorange},thick] coordinates {(0,1.063571) (1,1.800646) (2,3.672994)};\addlegendentry{Batch 4}
\end{axis}
\begin{axis}[scale only axis,width=7cm,height=1.65cm,xmin=-.1,xmax=4.1,ymin=.8,ymax=3.2,ytick={1,2,3},title={(b) Whole-model decoding},xlabel={Context length},xtick={0,1,2,3,4},xticklabels={32K,64K,128K,256K,512K},ylabel={Speedup ($\times$)},title style={font=\small},tick label style={font=\scriptsize},label style={font=\scriptsize},tick align=inside,grid=major,grid style={gray!15},legend style={font=\scriptsize,at={(.02,.98)},anchor=north west,draw=none,fill=white,fill opacity=.85,text opacity=1,row sep=-2pt},mark size=1.6pt]
\addplot[gray,dashed,forget plot] coordinates {(0,1)(4,1)};
\addplot[color=nfablue,mark=*,mark options={solid,fill=nfablue},thick] coordinates {(0,0.961631) (1,1.054425) (2,1.292574) (3,1.773728) (4,2.971175)};\addlegendentry{Batch 1}
\addplot[color=nfateal,mark=square*,mark options={solid,fill=nfateal},thick] coordinates {(0,0.988327) (1,1.189987) (2,1.608654) (3,2.745160)};\addlegendentry{Batch 2}
\addplot[color=nfaorange,mark=triangle*,mark options={solid,fill=nfaorange},thick] coordinates {(0,1.114747) (1,1.471229) (2,2.350404)};\addlegendentry{Batch 4}
\end{axis}
\end{tikzpicture}
\caption{NFA speedup over FlashInfer: (a) operator time including selection; (b) whole-model decoding. Dashed lines denote parity. Missing points exceed GPU memory capacity.}
\label{fig:decode}
\end{figure}
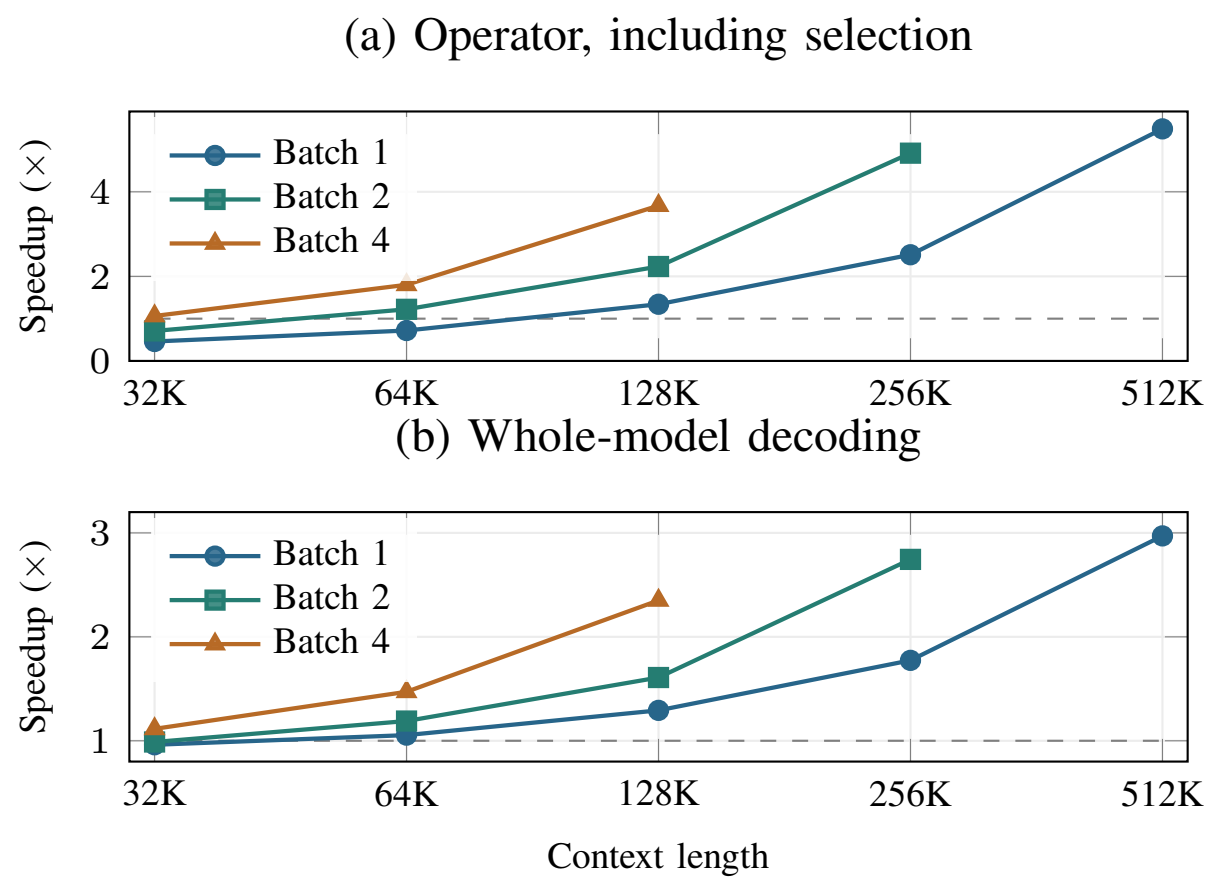

\textbf{Comparison with FlashInfer.} NFA's speedup over FlashInfer increases with context length and batch size (Fig.~\ref{fig:decode}). At 512K and BS1, NFA reaches $2.97\times$ whole-model speedup (b) and $5.49\times$ operator speedup including selection (a). At BS1, operator speedup exceeds parity from 128K, where attention savings outweigh selection costs. At 32K, whole-model decoding remains slightly slower for batches one and two.

\definecolor{nfagain}{HTML}{D5A51B}
\textbf{Comparison with sparse baselines.} Figure~\ref{fig:baseline-decode} compares whole-model decoding latency at batch size one. In this evaluation, NFA achieves the lowest latency across all tested context lengths, providing $1.25$--$3.97\times$, $1.81$--$5.14\times$, and $2.29$--$2.37\times$ speedups over JSSA, Quest, and AsyncTLS, respectively. NFA maintains nearly constant TPOT as context length grows. In the evaluated inputs, near-region skipping increases from 13 layers at 32K to 30 layers at 256K, helping offset the additional selection work at longer contexts.

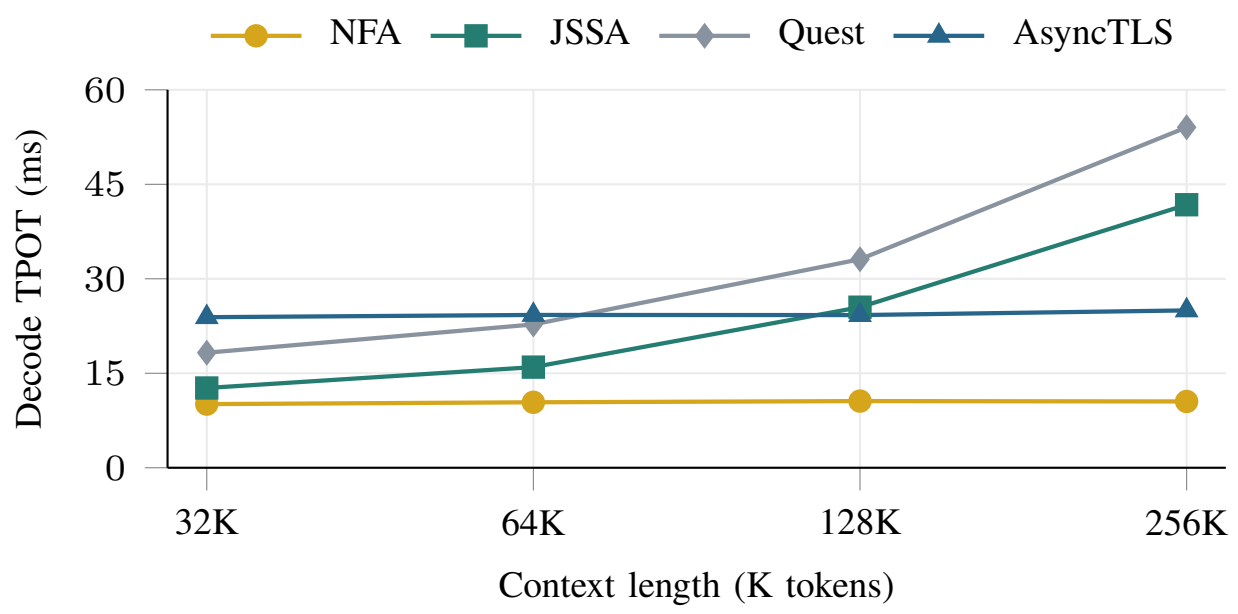
\begin{figure}[!htbp]
\centering
\begin{tikzpicture}
\begin{axis}[scale only axis,width=7cm,height=2.5cm,xmin=-.12,xmax=3.12,ymin=0,ymax=60,
xtick={0,1,2,3},xticklabels={32K,64K,128K,256K},ytick={0,15,30,45,60},
xlabel={Context length (K tokens)},ylabel={Decode TPOT (ms)},
tick label style={font=\scriptsize},label style={font=\scriptsize},
axis x line*=bottom,axis y line*=left,tick align=outside,grid=major,grid style={gray!16},
legend style={font=\scriptsize,at={(.5,1.04)},anchor=south,draw=none,fill=none,legend columns=4,column sep=3pt},mark size=1.8pt]
\addplot[thick,color=nfagain,mark=*] coordinates {(0,10.103648) (1,10.394048) (2,10.583296) (3,10.525536)};\addlegendentry{NFA}
\addplot[thick,color=nfateal,mark=square*] coordinates {(0,12.662208) (1,15.975040) (2,25.490208) (3,41.757118)};\addlegendentry{JSSA}
\addplot[thick,color=nfabaseline,mark=diamond*] coordinates {(0,18.262527) (1,22.753759) (2,33.100449) (3,54.050816)};\addlegendentry{Quest}
\addplot[thick,color=nfablue,mark=triangle*] coordinates {(0,23.903744) (1,24.259008) (2,24.239296) (3,24.980255)};\addlegendentry{AsyncTLS}
\end{axis}
\end{tikzpicture}
\caption{Whole-model decoding latency of NFA, JSSA, Quest, and AsyncTLS on Qwen3-8B at batch size one with a total attention budget of 1024 tokens, including protected positions. Latency includes selection overhead. Lower is better.}
\label{fig:baseline-decode}
\end{figure}

\subsection{Ablation Study}
\setlength{\intextsep}{5pt}\label{sec:ablation}
We evaluate each component with other settings fixed within each comparison; operator studies use Qwen3-8B at BS1.

\begin{figure}[!htbp]
\centering
\includegraphics[width=\columnwidth]{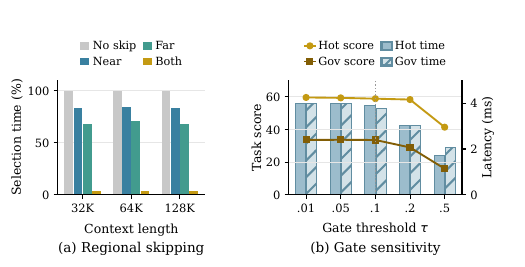}
\caption{Gate ablation on Qwen3-8B. (a) Selection latency under forced skip states, normalized to no skip. (b) Threshold sensitivity at $K_2=1024$. Lines: HotpotQA F1 (Hot) and GovReport ROUGE-L (Gov), 200 examples each. Bars: all-layer L1+L2+attention latency per decode step on one 32K input per task at BS1. The dotted line marks the default $\tau=0.1$.}
\label{fig:gate-ablation}
\end{figure}

\emph{Lightweight Gate Estimator.} We evaluate the savings from regional skipping and the quality--latency trade-off of the gate threshold (Fig.~\ref{fig:gate-ablation}). With 32D scoring fixed, skipping the near or far region accelerates selection by $1.19$--$1.20\times$ or $1.41$--$1.48\times$, respectively; skipping both reduces selection latency by approximately 96\% (a). We further vary $\tau$ on HotpotQA and GovReport with $K_2=1024$ (b). Scores change by at most 0.74 points over $\tau=0.01$--$0.1$, while larger thresholds trade quality for lower latency. We therefore use $\tau=0.1$ as the default. Gate estimation adds less than 1\% to dense-prefill time.

\begin{figure}[H]
\centering
\definecolor{nfagain}{HTML}{D5A51B}
\pgfplotsset{nfaablation/.style={scale only axis,width=3.05cm,height=1.5cm,tick label style={font=\scriptsize},label style={font=\scriptsize},title style={font=\scriptsize,yshift=20pt},axis x line*=bottom,axis y line*=left,axis line style={gray!65},tick align=outside,ymajorgrids,grid style={gray!16},legend style={font=\scriptsize,draw=none,fill=none,at={(.5,1.02)},anchor=south,column sep=2pt}}}
\begin{tikzpicture}
\begin{axis}[nfaablation,at={(.65cm,0)},anchor=south west,width=2.35cm,height=1.65cm,
xmin=0,xmax=62,xtick={0,20,40,60},ymin=-.5,ymax=4.5,
ytick={0,1,2,3,4},yticklabels={Random1,Random0,Uniform32,Last32,FullKV},
xlabel={Average score},title={(a) GLM task quality},ymajorgrids=false,xmajorgrids,clip=false,xbar,bar width=5pt,bar shift=0pt]

\addplot[draw=gray!55,fill=gray!40] coordinates {(47.96,4)};
\addplot[draw=nfagain!80!black,fill=nfagain] coordinates {(46.745,3)(47.435,2)(47.24,1)(47.645,0)};
\node[font=\scriptsize,anchor=west] at (axis cs:49.2,4) {47.96};
\node[font=\scriptsize,anchor=west] at (axis cs:48.0,3) {46.75};
\node[font=\scriptsize,anchor=west] at (axis cs:48.7,2) {47.44};
\node[font=\scriptsize,anchor=west] at (axis cs:48.5,1) {47.24};
\node[font=\scriptsize,anchor=west] at (axis cs:48.9,0) {47.65};
\end{axis}
\begin{axis}[nfaablation,at={(4.15cm,0)},anchor=south west,xmin=-.5,xmax=1.5,ymin=0,ymax=5.8,ytick={0,2,4},xtick={0,1},xticklabels={L1,L2},ylabel={Latency (ms)},title={(b) Qwen scoring},ybar,bar width=9pt,legend columns=2]
\addplot[draw=gray!55,fill=gray!40] coordinates {(0,1.206)(1,4.308)};\addlegendentry{128D}
\addplot[draw=nfagain!80!black,fill=nfagain] coordinates {(0,.878)(1,2.566)};\addlegendentry{32D}
\node[font=\scriptsize,text=nfagain!65!black,anchor=south] at (axis cs:0,1.5) {$1.37\times$};
\node[font=\scriptsize,text=nfagain!65!black,anchor=south] at (axis cs:1,4.6) {$1.68\times$};
\end{axis}
\end{tikzpicture}
\caption{Subspace ablation. (a) GLM average over HotpotQA/GovReport (200 examples each); all subsets use 32 NoPE coordinates. (b) Qwen3-8B scoring latency at 64K, gating disabled.}
\label{fig:subspace-ablation}
\end{figure}
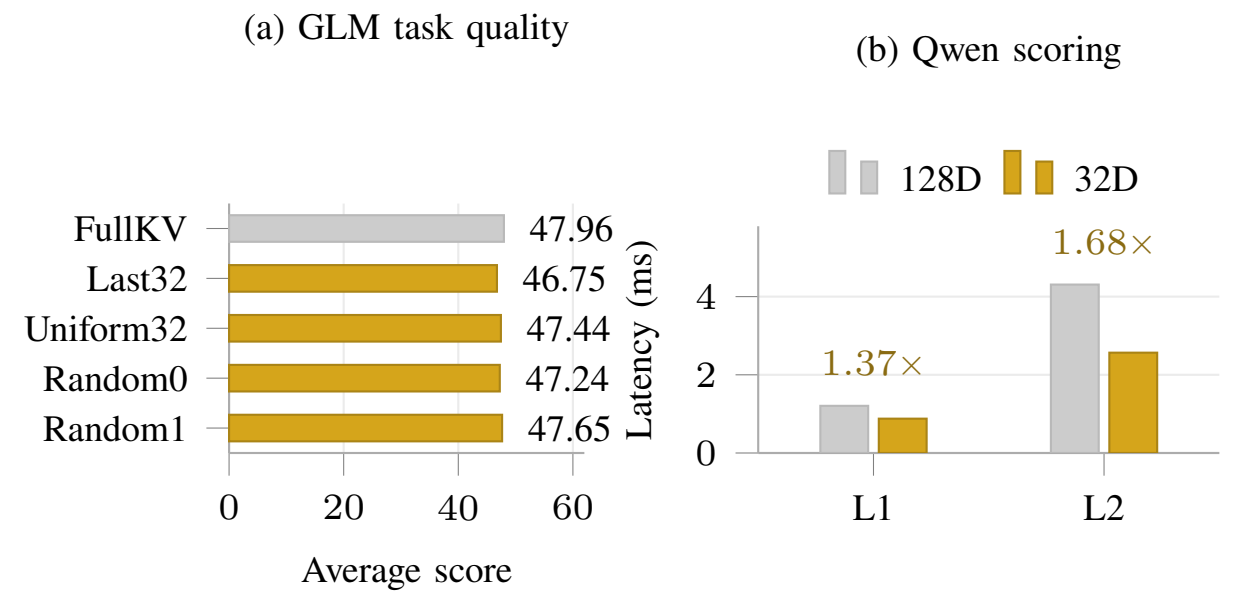

\emph{Calibration-Free Subspace Extractor.} We assess NoPE coordinate sensitivity on GLM and scoring cost on Qwen3-8B (Fig.~\ref{fig:subspace-ablation}). GLM’s tested 32D subsets retain comparable average quality to FullKV on HotpotQA/GovReport (a). On Qwen3-8B at 64K, 32D scoring accelerates L1 and L2 by $1.37\times$ and $1.68\times$, respectively, over full-dimensional (128D) scoring (b).

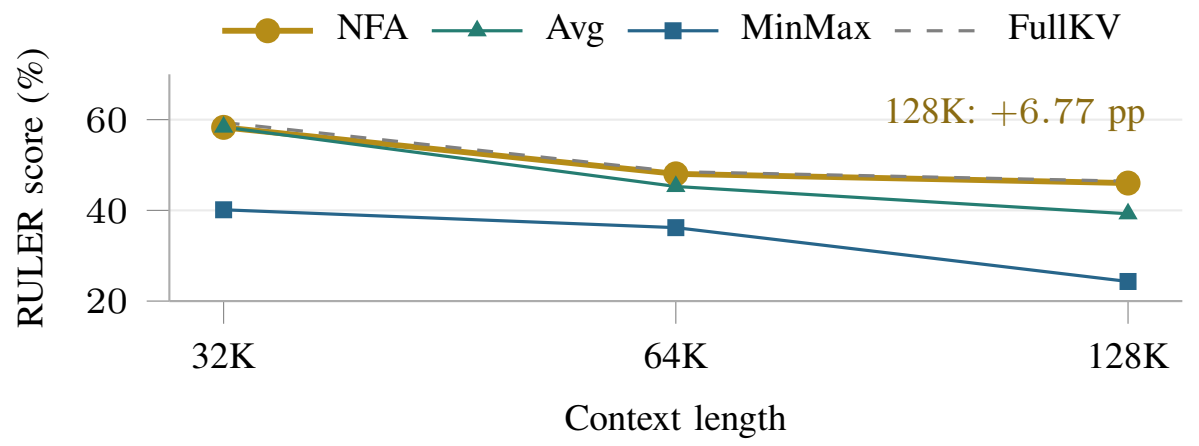
\begin{figure}[H]
\centering
\definecolor{nfagain}{HTML}{D5A51B}
\pgfplotsset{nfaablation/.style={scale only axis,width=3.05cm,height=1.5cm,tick label style={font=\scriptsize},label style={font=\scriptsize},title style={font=\scriptsize,yshift=20pt},axis x line*=bottom,axis y line*=left,axis line style={gray!65},tick align=outside,ymajorgrids,grid style={gray!16},legend style={font=\scriptsize,draw=none,fill=none,at={(.5,1.02)},anchor=south,column sep=2pt}}}
\begin{tikzpicture}
\begin{axis}[nfaablation,width=6.7cm,
xmin=-.12,xmax=2.12,ymin=20,ymax=70,ytick={20,40,60},
xtick={0,1,2},xticklabels={32K,64K,128K},
ylabel={RULER score (\%)},xlabel={Context length},legend columns=4]
\addplot[color=nfagain!85!black,line width=1.1pt,mark=*,mark size=1.8pt] coordinates {(0,58.30)(1,48.05)(2,46.02)};\addlegendentry{NFA}
\addplot[color=nfateal,semithick,mark=triangle*,mark size=1.5pt] coordinates {(0,58.37)(1,45.27)(2,39.25)};\addlegendentry{Avg}
\addplot[color=nfablue,semithick,mark=square*,mark size=1.3pt] coordinates {(0,40.13)(1,36.19)(2,24.32)};\addlegendentry{MinMax}
\addplot[color=gray,dashed,semithick] coordinates {(0,59.38)(1,48.54)(2,46.43)};\addlegendentry{FullKV}
\node[font=\scriptsize,text=nfagain!65!black,anchor=north east] at (axis cs:2.08,69) {128K: $+6.77$ pp};
\end{axis}
\end{tikzpicture}
\caption{NFA regional page selection versus single-pool Avg and MinMax on Qwen3-8B across RULER context lengths.}
\label{fig:regional-ablation}
\end{figure}

\emph{Distance-Aware Two-Level Selector.} To assess regional selection, we compare NFA with single-pool Avg and MinMax on RULER, holding the total page and token budgets, scoring subspace, and L2 selection fixed (Fig.~\ref{fig:regional-ablation}). At 128K, NFA gains 6.77 and 21.70 points, respectively.

\section{Conclusion}
We presented NFA, which combines input-adaptive regional
skipping, shared scoring subspaces, and near--far
selection for efficient long-context decoding.
Evaluations demonstrate decoding speedups over FlashInfer
while maintaining aggregate task quality comparable
to full attention.

\raggedbottom
\clearpage
\balance


\begin{thebibliography}{99}
\bibitem{deepseekv41} DeepSeek-AI, ``DeepSeek-V4.1-Flash: Pushing the Limits of KV Cache Compression,'' arXiv preprint arXiv:2609.19969, 2026.
\bibitem{longcatlsa} W. Zan \emph{et al.}, ``LongCat Sparse Attention: Taming the Lightning via Streaming-aware Hierarchical Cross-Layer Indexing,'' arXiv preprint arXiv:2608.01662, 2026.
\bibitem{quest} J. Tang, Y. Zhao, K. Zhu, G. Xiao, B. Kasikci, and S. Han, ``Quest: Query-Aware Sparsity for Efficient Long-Context LLM Inference,'' in \emph{Proc. ICML}, 2024.
\bibitem{hisa} Y. Xu \emph{et al.}, ``HISA: Efficient Hierarchical Indexing for Fine-Grained Sparse Attention,'' in \emph{Proc. COLM}, 2026.
\bibitem{jssa} J. Song, Y. Kim, and J.-J. Kim, ``Joint Subspace Sparse Attention for Efficient Long-Context Decoding,'' arXiv preprint arXiv:2610.19026, 2026.
\bibitem{asynctls} Y. Hu \emph{et al.}, ``AsyncTLS: Efficient Generative LLM Inference with Asynchronous Two-level Sparse Attention,'' arXiv preprint arXiv:2604.07815, 2026.
\bibitem{gqa} J. Ainslie \emph{et al.}, ``GQA: Training Generalized Multi-Query Transformer Models from Multi-Head Checkpoints,'' in \emph{Proc. EMNLP}, 2023.
\bibitem{rope} J. Su, Y. Lu, S. Pan, A. Murtadha, B. Wen, and Y. Liu, ``RoFormer: Enhanced Transformer with Rotary Position Embedding,'' arXiv preprint arXiv:2104.09864, 2021.
\bibitem{deepseekv2} DeepSeek-AI \emph{et al.}, ``DeepSeek-V2: A Strong, Economical, and Efficient Mixture-of-Experts Language Model,'' arXiv preprint arXiv:2405.04434, 2024.
\bibitem{fasa} Y. Wang \emph{et al.}, ``FASA: Frequency-Aware Sparse Attention,'' in \emph{Proc. ICLR}, 2026.
\bibitem{clusterkv} G. Liu \emph{et al.}, ``ClusterKV: Manipulating LLM KV Cache in Semantic Space for Recallable Compression,'' in \emph{Proc. DAC}, 2025, doi: 10.1109/DAC63849.2025.11132479.
\bibitem{streaming} G. Xiao, Y. Tian, B. Chen, S. Han, and M. Lewis, ``Efficient Streaming Language Models with Attention Sinks,'' in \emph{Proc. ICLR}, 2024.
\bibitem{duoattention} G. Xiao \emph{et al.}, ``DuoAttention: Efficient Long-Context LLM Inference with Retrieval and Streaming Heads,'' in \emph{Proc. ICLR}, 2025.
\bibitem{fastgen} S. Ge, Y. Zhang, L. Liu, M. Zhang, J. Han, and J. Gao, ``Model Tells You What to Discard: Adaptive KV Cache Compression for LLMs,'' in \emph{Proc. ICLR}, 2024.
\bibitem{qwen3} A. Yang \emph{et al.}, ``Qwen3 Technical Report,'' arXiv preprint arXiv:2505.09388, 2025.
\bibitem{llama3} A. Grattafiori \emph{et al.}, ``The Llama 3 Herd of Models,'' arXiv preprint arXiv:2407.21783, 2024.
\bibitem{glm47flash} Z.ai, ``GLM-4.7-Flash,'' model card. [Online]. Available: \url{https://huggingface.co/zai-org/GLM-4.7-Flash}. Accessed: Nov. 16, 2026.
\bibitem{longbench} Y. Bai \emph{et al.}, ``LongBench: A Bilingual, Multitask Benchmark for Long Context Understanding,'' in \emph{Proc. ACL}, 2024.
\bibitem{longbenchv2} Y. Bai \emph{et al.}, ``LongBench v2: Towards Deeper Understanding and Reasoning on Realistic Long-context Multitasks,'' in \emph{Proc. ACL}, pp. 3639--3664, 2025.
\bibitem{ruler} C.-P. Hsieh \emph{et al.}, ``RULER: What's the Real Context Size of Your Long-Context Language Models?,'' in \emph{Proc. COLM}, 2024.
\bibitem{flashinfer} Z. Ye \emph{et al.}, ``FlashInfer: Efficient and Customizable Attention Engine for LLM Inference Serving,'' in \emph{Proc. MLSys}, 2025.
\bibitem{yarn} B. Peng, J. Quesnelle, H. Fan, and E. Shippole, ``YaRN: Efficient Context Window Extension of Large Language Models,'' arXiv preprint arXiv:2309.00071, 2023.
\end{thebibliography}
\end{document}